\documentclass[letterpaper]{article}
\usepackage{proceed2e}
\usepackage[margin=1in]{geometry}
\usepackage{times}
\usepackage{amsmath,amsfonts,amssymb,mathtools,mathrsfs,dsfont,bm}
\usepackage{amsthm}
\usepackage{graphicx}
\usepackage{subcaption}
\usepackage{wrapfig}
\usepackage{hyperref}
\usepackage[normalem]{ulem}
\usepackage{enumitem}
\usepackage{natbib}
\usepackage{dblfloatfix}
\usepackage{float}
\usepackage{multicol}
\usepackage{multirow}
\usepackage{array}
\usepackage{tabu}
\usepackage{hhline}
\usepackage{ragged2e}
\usepackage[dvipsnames]{xcolor}
\usepackage{colortbl}
\usepackage{rotating}
\usepackage[linesnumbered,ruled,vlined,algo2e]{algorithm2e}
\usepackage{tikz}
\usepackage{afterpage}

\newlength{\oldtextfloatsep}

\SetCommentSty{codecommentstyle}

\newcommand\Tstrut{\rule{0pt}{2.0ex}}       
\newcommand\Bstrut{\rule[-0.6ex]{0pt}{0pt}}	
\newcommand{\TBstrut}{\Tstrut\Bstrut} 		

\definecolor{pale_gray}{rgb}{0.95,0.95,0.95}
\definecolor{pale_blue}{rgb}{0.90,0.95,1}
\newcommand{\best}[1]{\textcolor{red}{#1}}
\newcommand{\side}[1]{\begin{sideways}\hspace{-1.3em}#1\end{sideways}}

\title{Agreement-based Learning}

\author{
{\bf Emmanouil Antonios Platanios} \\
Machine Learning Department \\
Carnegie Mellon University \\
Pittsburgh, PA 15213}

\begin{document}

\maketitle

\begin{abstract}

Model selection is a problem that has occupied machine learning researchers for a long time. Recently, its importance has become evident through applications in deep learning. We propose an agreement-based learning framework that prevents many of the pitfalls associated with model selection. It relies on coupling the training of multiple models by encouraging them to agree on their predictions while training. In contrast with other model selection and combination approaches used in machine learning, the proposed framework is inspired by human learning. We also propose a learning algorithm defined within this framework which manages to significantly outperform alternatives in practice, and whose performance improves further with the availability of unlabeled data. Finally, we describe a number of potential directions for developing more flexible agreement-based learning algorithms.

\end{abstract}

\section{INTRODUCTION}

Model selection is a problem that traces itself back to the 14th century when William of Ockham argued that among competing hypotheses, the one with the fewest assumptions should be selected. This principle is known as {\em Occam's razor} and centuries later, model selection often occupies the minds of machine learning researchers. While developing algorithms that learn patterns, researchers create models that represent the problem being solved. There often exists a large set of promising models and the researcher has to decide which one to use. There exist various methods for making that decision, with cross-validation \citep{Kohavi:1995} being one of the most frequently used in practice. However, researchers often have to rely on their intuition and expertise, and spend a significant amount of time fine-tuning their models. This is especially true for {\em deep learning} which is arguably the most popular area of research in machine learning, at present. We propose a new learning framework, {\em agreement-based learning}, which prevents many of the pitfalls associated with model selection. It relies on coupling the training of multiple models and combining their predictions into a single output. This coupling is performed by encouraging the models to agree with each other on their predictions, while training. We show how the proposed framework is inspired by human learning, in contrast with other approaches commonly used in machine learning. Finally, we provide experimental results which exhibit how our framework successfully manages to outperform cross-validation and other ensemble methods that do not couple the training of the models.

\begin{figure}[t!]
	\centering
    \includegraphics[width=0.5\textwidth, trim=9.2cm 2.5cm 1.5cm 6cm, clip]{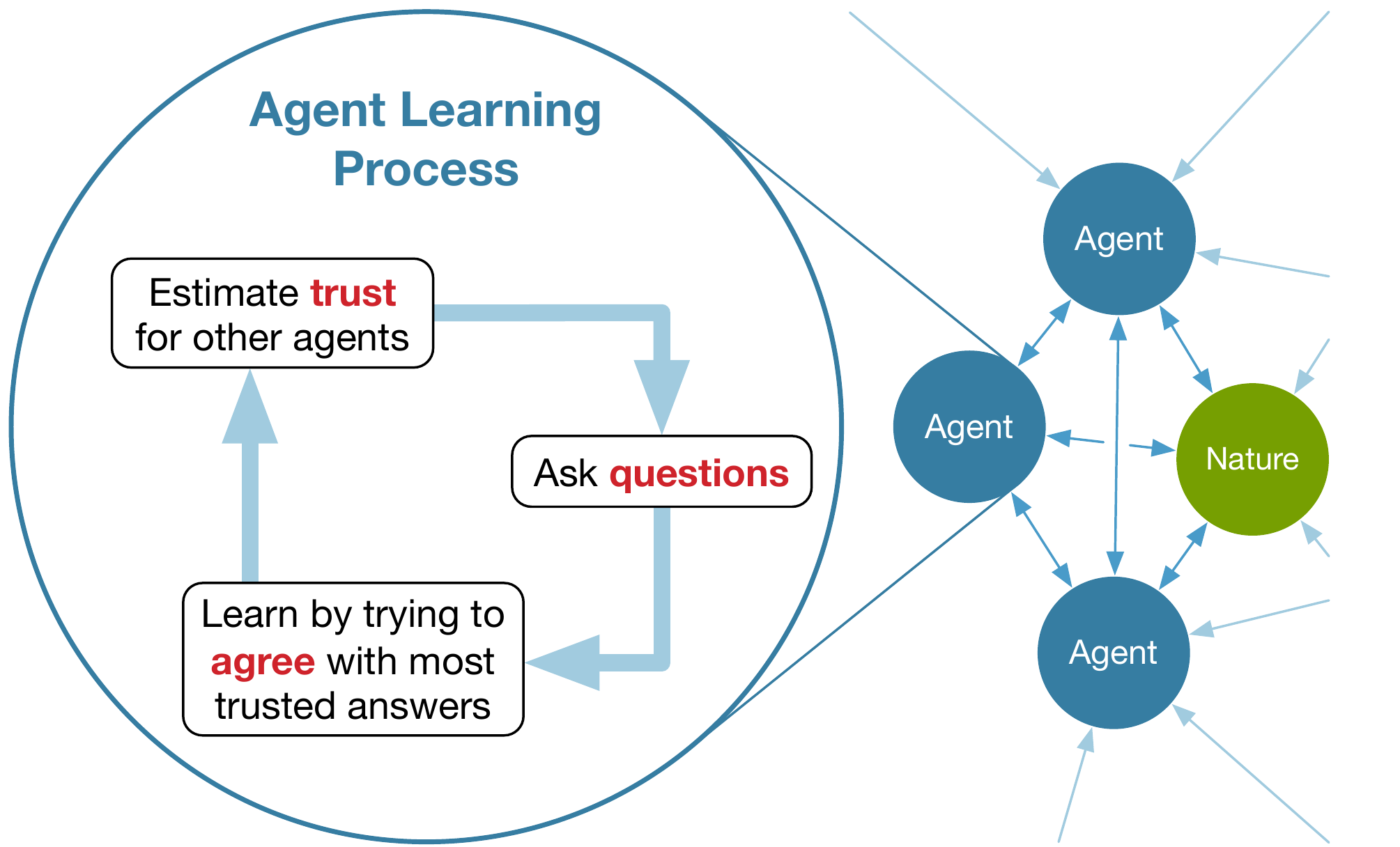}
	\caption{Illustration of the agreement-based learning framework that we propose.}
	\vspace{-1.0em}
	\label{fig:agreement_based_learning_illustration}
\end{figure}

We start by discussing the relationship between machine learning and human learning. Then, we propose a model for natural learning that resembles what happens in the real-world and forms the underlying idea of our agreement-based learning framework, which is formally introduced in section \ref{sec:agreement_based_learning}. In section \ref{sec:related_work}, we discuss existing work in this direction and how they differ from our proposal. Finally, in \ref{sec:experiments} we provide an extensive experimental evaluation of the agreement-based learning framework.

\subsection{MACHINE VS NATURAL LEARNING}
\label{sec:machine_vs_natural_learning}

{\em Supervised learning} is one of the first and still dominant approaches used to make machines able to learn. It comprises providing the machine a set of input-output pairs and expecting it to learn a {\em generalized} mapping between inputs and outputs. Thus, the machine must learn to produce correct outputs for new --- previously unseen --- inputs. This process is inspired by the way in which humans learn. Children ask people to name things they do not recognize. For example, they will point to a mug and ask what that object is. After having seen a few examples they become able to recognize mugs, even if they look different than the mugs they have seen before. However, when people ask questions, the answers are often provided by other humans (as opposed to the surrounding natural environment, for example) and may not always be correct. Furthermore, due to the limited answers that the surrounding natural environment can provide, much of the human ability to generalize could be attributed to the interaction with other humans and this kind of question-answering.

These observations inspired us to view learning from a different perspective. Let us refer to each human and animal in nature as an {\em agent}. And let us do the same for nature itself (e.g., nature is an agent which tells other agents that if they drop an apple from a building, it will fall down). We argue that learning can be defined as a process of interaction between these agents, through which they try to agree with each other. More specifically, we could argue that each agent: (i) decides how much to {\em trust} other agents, (ii) asks  multiple other agents a question\footnote{``Asking a question'' is quite an abstract notion here. For example, dropping an apple from a building and observing what happens could constitute a question for the nature agent.}, and (iii) forms an idea as to what the true answer to the question may be by weighting the answers based on how much the agents that provided them are trusted. Learning can thus be defined as a process in which agents learn by trying to agree with each other.

In machine learning, ``agents'' can refer to different learning algorithms\footnote{By learning algorithm we refer to the algorithm and its specific underlying model. For models that have a set of hyperparameters, different values of these hyperparameters result in entirely different learning algorithms, and thus different agents.} and instead of training each one independently, this model of learning proposes that their training procedures are coupled via the constraint that, while training, the agents try to agree in their predictions. Furthermore, input-output pairs that are known to be true, as traditionally used in supervised learning, can still be provided through an agent that is fixed (i.e., does not learn, such as the previously introduced nature agent). This model of learning is the underlying idea and main motivation for the {\em agreement-based learning} framework that we propose in this paper. We believe it can improve upon the supervised learning model in multiple ways:
\begin{enumerate}[noitemsep, topsep=0pt, leftmargin=*]
	\item \uline{Sharing Information:} Different agents may have access to different information. For humans, this could be due to growing up in different environments whereas for learning algorithms, it could be due to the fact that different models rely on and incorporate different kinds of information (e.g., some models might use text data as input whereas some others might use image data). Agreement-based learning allows agents to implicitly share the different information they may have by providing answers to questions that other agents cannot.
	\item \uline{Preventing Overfitting:} {\em Overfitting} is a well-studied concept in machine learning. When an algorithm uses a complex model and there is limited training data, the algorithm may learn a mapping function from inputs to outputs that is much more complex than the true underlying function. This would result in bad generalization and potentially many wrong predictions for inputs the algorithm has not seen during its training. For example, modeling a linear function using a tenth order polynomial function while having observed only 3 points of the underlying line, would likely result in a bad fit. The same is true for humans to some extent. People tend to ``overfit'' in their own beliefs due to the limited experience  they might have and the bias of only acknowledging observations that confirm their earlier beliefs (i.e., {\em confirmation bias}). Agreement-based learning can help prevent overfitting since it is less likely that multiple agents would overfit and agree simultaneously. This is especially true if they have access to different information.
	\item \uline{Using Unlabeled Data:} In machine learning, we refer to the provided input-output pairs as {\em labeled data} (since each input is ``labeled'' with some output). The agreement-based learning framework makes agents capable of also using {\em unlabeled data}, which consist of inputs without the corresponding outputs. This is due to the fact that agents can strive to make their outputs for these inputs agree, irrespective of the fact that no true outputs are provided. The framework that we propose can thus also be used to convert any supervised learning algorithm to a {\em semi-supervised learning} algorithm. This will become more clear in section \ref{sec:agreement_based_learning}, when we formally introduce our proposed approach.
\end{enumerate}

Note that there exist situations in which learning can fail in this context; for example, all agents may agree on something that is wrong. Such situations challenge the notion of what is the truth that the agents are trying to learn, and what truth is, more generally. These questions are of a philosophical nature and are beyond the scope of this paper. However, in contrast to traditional machine learning methods, our approach does not assume an inherently true underlying distribution of data and can thus work in settings in which truth is relative. In the setting that we consider, there always exists a {\em nature agent} that provides answers to some questions and is fixed (i.e., does not learn and thus does not change its answers to questions with time). Since there cannot be total agreement unless other agents agree with the nature agent, the degenerate case described above is avoided.

\section{RELATED WORK}
\label{sec:related_work}

The literature covers many projects that are either explicitly or implicitly related to agreement-based learning. A related area of research within the broad spectrum of machine learning is that of ensemble methods \citep{Dietterich:2000}. These are methods that take as input the outputs of multiple models, or in some cases generate these models themselves, and combine them into a single output. The simplest ensemble method is majority voting where the most popular value among the outputs of the models is selected as the combined output. Bagging \citep{Breiman:1996b} and boosting \citep{Breiman:1996a} are methods that, instead of just taking the outputs of multiple models and combining them, they also couple their training. They do so by manipulating the distribution of the data provided to the models while training. \citet{Freund:1997} proposed one of the most influential and widely used boosting algorithms, known as AdaBoost. AdaBoost changes the distribution of the training data such that the training algorithm can focus more on examples for which it has made erroneous predictions in the past. However, in contrast to the framework that we propose, AdaBoost places constraints on the functional form of the ensemble models (i.e., their predictions have to be boolean-valued scalars). Furthermore, bagging and boosting algorithms do not make use of unlabeled data that may be available during training.

Researchers have used the concept of agreement between multiple models in other ways, too. For example, \citet{Hanneke:2014} reviews methods that use disagreement between models to perform active learning, and others have used some notion of agreement between models in order to estimate their error rates \citep{Collins:1999, Dasgupta:2001, Bengio:2003, Madani:2004, Schuurmans:2006, Balcan:2013, Parisi:2014, Platanios:2014, Platanios:2016, Anonymous:logic}. The work most relevant to our proposal, though, is co-training \citep{Blum:1998}. In co-training, a set of models is trained using the following iterative process: (i) each model is trained independently given a small set of initial training data, (ii) the models make predictions on a set of unlabeled data, (iii) their most confident predictions are added to the training dataset, and (iv) the process repeats itself from step (i). Note that, even though the models do not directly attempt to maximize agreement between themselves, they implicitly do so. This is due to the fact that, through the common training dataset that they can append their predictions to, the models effectively train each other. One possible pitfall of co-training is that if a model always produces confident yet wrong predictions, it can negatively affect the performance of other models by corrupting the training dataset. This pitfall is avoided with our framework since the models are not necessarily blindly trusted and their predictions are not equally weighted.

\section{PROPOSED METHOD}

We consider a traditional machine learning setting in which we have a set of models\footnote{Note that by ``different,'' here, we do not only refer to structural differences in the models, but we include differences in hyperparameters among instantiations of the same model. For example, in the context of deep learning one could set $f_1, \hdots, f_M$ to be multi-layer perceptrons (MLPs) with different architectures or activation functions, or even completely different neural networks.}, $f_1, \hdots, f_M$, that, given some input $x$, produce some output $f_j(x)$. We assume that these models have parameters that are learned by minimizing a loss function of the form:
\begin{equation}
\label{eq:loss_function}
	\mathcal{L}_j\left(\{x_i, y_i\}_{i=1}^N\right) = \sum_{i=1}^N{\ell_j(f_j(x_i), y_i)},
\end{equation}
where $j=1,\hdots,M$, $\{x_i, y_i\}_{i=1}^N$ is the set of training examples, and $\ell_j$ is the loss incurred for a single training example and it penalizes disagreement of the model prediction, $f_j(x_i)$, with the provided label, $y_i$ (e.g., cross-entropy for classification problems or mean squared error for regression problems). This function can be minimized using an appropriate optimization algorithm (e.g., gradient descent). This is a general supervised learning setting that covers both classification and regression problems.

\subsection{TRADITIONAL APPROACH}

Traditionally, a machine learning researcher would train all these different models separately and then do one of two things:
\begin{enumerate}[noitemsep, topsep=0pt, leftmargin=*]
	\item \uline{Model Selection:} Choose a single model for making predictions. A commonly used method to perform model selection is cross-validation \citep{Kohavi:1995}, mainly due to its strong theoretical guarantees. Selecting one model is what is frequently done in deep learning when researchers talk about {\em parameter tuning} (see e.g., the work of \citet{Bergstra:2012}).
	\item \uline{Model Combination (Ensemble):} Combine the predictions of all models into a single prediction using some kind of ensemble method \citep{Dietterich:2000}. A simple yet powerful and frequently used ensemble method is majority voting, where the predictions are combined by taking their mean.
\end{enumerate}

We propose and describe, in the remainder of this paper, a type of ensemble method in which the training of the models is coupled (and thus the models cannot be trained separately). However, in contrast to other model combination approaches that do this (e.g., boosting \citep{Breiman:1996a}), our approach does not impose any strict constraints on the functional form of the models, other than that their parameters can be learned by minimizing a loss function of the form presented in equation \ref{eq:loss_function}.

\subsection{AGREEMENT-BASED LEARNING}
\label{sec:agreement_based_learning}

Inspired by natural learning (section \ref{sec:machine_vs_natural_learning}), we propose a new approach for training, different than the traditional approaches described in the previous section. In order to consider agreement between models during training, we define an {\em augmented loss function} that replaces the loss function of equation \ref{eq:loss_function}:
\begin{equation}
\label{eq:augmented_loss_function}
\begin{aligned}
	&\mathcal{AL}_j\left(\{x_i, y_i\}_{i=1}^N, \{x_i'\}_{i=1}^{N'}\right) = \\
	&\qquad\;\mathcal{L}_j\left(\{x_i, y_i\}_{i=1}^N\right) + \lambda_j \sum_{i=1}^{N'}{\ell_j'(f_j(x_i'), \hat{f}(x_i'))},
\end{aligned}
\end{equation}
where $\{x_i'\}_{i=1}^{N'}$ is a  set of training examples that are unlabeled, $\hat{f}$ represents a combined prediction from all models, which we shall call the {\em consensus prediction} and which is defined in the next section, $\ell_j'$ is a loss function that penalizes disagreement of the model prediction with the consensus prediction, and $\lambda_j \geq 0$ is a non-negative number which represents the tendency of model $j$ to agree with the consensus. For simplicity, we set $\ell_j'=\ell_j$ and $\lambda_j=1$ for all $j=1,\hdots,M$, even though our framework supports any non-negative value\footnote{Note that, in principle, negative values can also be used, but this would entail that an agent is trying to disagree with the other agents. This is an interesting idea to explore, but is outside the scope of this paper.}. It remains to: (i) define the consensus prediction (section \ref{sec:consensus}), (ii) define the training algorithm that minimizes the augmented loss function (section \ref{sec:coupling_training}), (iii) define the way in which predictions are made (section \ref{sec:predictions}), and (iv) point out the relationship between this augmented loss function and the ideas presented in section \ref{sec:machine_vs_natural_learning}, regarding natural learning (section \ref{sec:relationship_to_natural_learning}).

This augmented loss function effectively forces the model to not only try and agree with the provided input-output pairs, but to also --- at the same time --- try and agree with the other models on other inputs for which we do not know the corresponding outputs. This equation also makes it easy to see why our proposed approach helps prevent overfitting, as mentioned in section \ref{sec:machine_vs_natural_learning}. The added term to the loss function can be interpreted as a regularizer to the original optimization problem. It is thus easy to see why the added term can help the optimization solver of the original problem avoid local minima and find a better solution.


\subsubsection{Consensus Definition}
\label{sec:consensus}

The consensus prediction, $\hat{f}(x)$, combines the predictions of all models, $f_1(x),\hdots,f_M(x)$, into a single prediction. It could be generated using any existing ensemble method \citep{Dietterich:2000}. For ensemble methods that require no training, such as majority voting, the definition of the consensus prediction is straightforward. However, some consensus methods, including the ones described in the rest of this section, require training. Section \ref{sec:coupling_training} describes the exact way in which they are trained and in which the training procedures of all models are coupled together. In the rest of this section, we describe the consensus methods that we used for our experiments. In our experiments we consider multi-label classification problems and thus, we henceforth assume that the model predictions are vectors containing probabilities (i.e., each element corresponds to the probability of a particular label being positive).

\paragraph{Trainable Majority Vote.} The majority vote consensus is defined as the mean of the model predictions:
\begin{equation}
	\hat{f}_{MV}(x) \triangleq \frac{1}{M}\sum_{j=1}^M{f_j(x)}.
\end{equation}
A simple extension to this consensus method is to consider a weighted combination of the model predictions:
\begin{equation}
	\hat{f}_{TMV}(x) \triangleq \frac{1}{M}\sum_{j=1}^M{w_jf_j(x)},\textrm{ where }\sum_{j=1}^M{w_j}=1,
\end{equation}
where the weights can be learned by minimizing a loss function. In the context of our framework, the loss function being minimized is the one defined in equation \ref{eq:loss_function}. For our experiments, we used the Adam optimizer \citep{Kingma:2014} to learn the values for the weights that minimize that objective function.

\paragraph{Semi-Supervised RBM Consensus.} This is a semi-supervised extension of the method proposed by \citet{Shaham:2016}. The original method that the authors propose is, to the extent of our knowledge, the current state-of-the-art for unsupervised ensemble learning, which is partly why we chose to use it, in addition to ease of implementation. The authors originally propose using a Restricted Boltzmann Machine (RBM) for unsupervised ensemble learning. In a multi-label classification setting, we define one RBM per label which uses the sigmoid activation function; the visible units of each RBM correspond to the model predictions, and there exists a single hidden unit that corresponds to the consensus prediction. More specifically in our setting, we have that:
\begin{equation}
\begin{aligned}
	&\mathbb{P}(\hat{f}(x), f_1(x), \dots, f_M(x)) = \\
	&\;\frac{1}{Z}\exp\!\bigg[a\hat{f}(x)\!+\!\sum_{j=1}^M{\!b_jf_j(x)}\!+\!\sum_{j=1}^M{\!\!w_jf_j(x)\hat{f}(x)}\bigg],
\end{aligned}
\end{equation}
where $\mathbb{P}$ denotes a probability, $a$, $b_j$, and $w_j$, for $j=1,\hdots,M$, are trainable parameters, and $Z$ is a normalizing constant to ensure the probability over all prediction values is valid (i.e., sums to $1$). The RBM is trained by maximizing the likelihood of the observed data using gradient-based optimization. The consensus prediction can then be computed by taking the mode of the following conditional distribution:
\begin{equation}
\begin{aligned}
	&\mathbb{P}(\hat{f}(x) = 1 \mid f_1(x),\dots,f_M(x)) = \\
	&\qquad\qquad\qquad\qquad\qquad\;\sigma\bigg[a + \sum_{j=1}^M{w_jf_j(x)}\bigg],
\end{aligned}
\end{equation}
where $\sigma(x)=\frac{1}{1+e^{-x}}$ is the sigmoid function.
Our extension of this RBM-based method also uses gradient descent to train the RBM. However, the likelihood function we consider consists of two terms: one using unlabeled data (which is the same as that used by \citet{Shaham:2016}), and one using labeled data, which is easy to define, given the definition of the RBM model using the sigmoid activation function.


Note that the agreement-based learning framework that we define is agnostic with respect to the chosen consensus method. This means that one could ``plug in'' any reasonably performing ensemble method and expect similar results. The choice of consensus method could depend on assumptions made about the models being trained. Some alternative methods that make different assumptions were proposed by \citet{Platanios:2014, Platanios:2016, Anonymous:logic}.

\subsubsection{Coupling the Training Procedures}
\label{sec:coupling_training}

The consensus prediction, $\hat{f}$, in equation \ref{eq:augmented_loss_function} is a function of all model predictions. This means that if we want to train our models using the augmented loss function, they all need to be trained in tandem. For this reason, we propose algorithm \ref{alg:agreement_based_learning}. The algorithm assumes that an iterative optimization algorithm is used to minimize the loss function of each model. This optimization algorithm is called in the parameter update steps of lines 10 and 12. For example, these steps could represent a gradient descent step. At every iteration of the optimization, the following steps are performed:
\begin{enumerate}[noitemsep, topsep=0pt, leftmargin=*]
	\item An unlabeled and a labeled data batch are sampled from the training data. The sizes of those batches can be chosen arbitrarily, but note that the relative sizes of each batch will have a balancing effect similar to $\lambda_j$ in equation \ref{eq:augmented_loss_function}. In our experiments, we set them equal.
	\item The consensus prediction is computed for the unlabeled data batch, using the current parameter values for each model.
	\item The parameters of each model are updated so that the value of the augmented loss function of equation \ref{eq:augmented_loss_function} is reduced. Note that for the first few iterations ($K_0$ in algorithm \ref{alg:agreement_based_learning}), the original model loss function is used instead; this is due to the fact that in the beginning of training, all models are expected to produce bad predictions and thus, forcing them to agree may not be a good idea. Instead, we can wait until they can perform better. Most consensus methods are only guaranteed to work well if the majority of the models produce predictions better than chance (i.e., random). A simple heuristic is to start using the augmented loss function after at least half of the models have performance on the training data higher than some threshold (e.g., when train accuracy exceeds $50\%$).
	\item The consensus method is re-trained every few iterations ($K$ in algorithm \ref{alg:agreement_based_learning}). Note that this re-training step could also be performed at every iteration but it is preferable not to do so, since: (i) the model parameters do not generally change dramatically between each iteration, and (ii) it may be prohibitively expensive to do so at every step.
\end{enumerate}

\setlength{\textfloatsep}{5pt}
\begin{algorithm2e}[!t]
\caption{Agreement-based learning algorithm.}
\label{alg:agreement_based_learning}
\small
\KwIn{Models $f_1,\hdots,f_M$, consensus method $\hat{f}$, unlabeled dataset $\mathcal{D}_U=\{x_i'\}_{i=1}^{N'}$, labeled dataset $\mathcal{D}_L=\{x_i, y_i\}_{i=1}^N$, unlabeled data batch size $N_U$, labeled data batch size $N_L$, consensus train burn-in iterations $K_0$, consensus retrain frequency $K$}
Initialize all model and consensus method parameters\\
$iteration \leftarrow 0$ \\
\While{at least one model has not converged}{
	Sample unlabeled data batch $\mathcal{B}_U$ of size $N_U$ from $\mathcal{D}_U$\\
	Sample labeled data batch $\mathcal{B}_L$ of size $N_L$ from $\mathcal{D}_L$\\
	Compute consensus prediction $\smash{\hat{f}(x)}$, for all $x\in\mathcal{B}_U$\\
	\For{$j=1,\hdots,M$}{
		\If{model $f_j$ has not converged}{
			\uIf{$iteration > K_0$}{
				Update model $f_j$ parameters so that the augmented loss, $\mathcal{AL}_j$, of equation \ref{eq:augmented_loss_function} decreases\\}
			\Else{
				Update model $f_j$ parameters so that the loss, $\mathcal{L}_j$, of equation \ref{eq:loss_function} decreases\\
			}
		}
	}
	\If{$iteration = 0 \textrm{\normalfont{ mod }} K$}{
		Re-train $\hat{f}$ using $\mathcal{D}_U$ and $\mathcal{D}_L$
	}
	$iteration \leftarrow iteration + 1$\\
}
\KwOut{Trained models $f_1,\hdots,f_M$ and trained consensus $\hat{f}$.}
\afterpage{\global\setlength{\textfloatsep}{\oldtextfloatsep}}
\end{algorithm2e}

\paragraph{Convergence.} From the definition of the augmented loss function in equation \ref{eq:augmented_loss_function} and algorithm \ref{alg:agreement_based_learning}, it is clear that, while training, the model parameters change and thus the function being optimized is not a fixed function. In general, this could mean that the iterative optimization procedure might not converge to a solution. However, we expect that as training proceeds, the model predictions will start agreeing more often\footnote{This is expected since minimizing the augmented loss function involves maximizing agreement through the consensus term.}. Thus, the augmented loss function will start behaving more like the original model loss function, retaining its convergence properties. Rather than theoretical convergence guarantees, we provide empirical evidence by noting that the algorithm converged in all of our experiments (presented in section \ref{sec:experiments}).

\paragraph{Scalability.} For the proposed algorithm we can only parallelize over the different model parameter updates (i.e., parallelize the ``for'' loop starting in line 7 of algorithm \ref{alg:agreement_based_learning}). This makes our method scalable, but we can do even better. Even though in our proposed algorithm, the optimization iterations of each model are synchronous, training could also be done in a completely asynchronous and distributed manner. This way, if one model's optimization iterations are faster than those of other models, it would not have to wait until all others have performed their iterations; it could instead proceed with its own optimization loop. In section \ref{sec:relationship_to_natural_learning} we discuss some extensions that would result in better scalability properties.

\paragraph{Performance.} An interesting observation regarding our proposed algorithm is that, as mentioned earlier, it can take a set of supervised models and train them in a semi-supervised manner. We expect this to result in better performance (e.g., better prediction accuracy) relative to a purely supervised approach, as the availability of unlabeled data increases. This is due to the fact that, as discussed in section \ref{sec:machine_vs_natural_learning}, the agreement-based learning framework can improve generalization and help prevent overfitting. Indeed, as shown in section \ref{sec:experiments}, our experiments confirm this expectation.

\subsubsection{Making Predictions}
\label{sec:predictions}

As per the definition of the consensus prediction in section \ref{sec:consensus}, it follows naturally to use the consensus prediction as the output prediction of our algorithm. This also means that the algorithm we propose is a type of ensemble method in which the training of the models is coupled.

Note that the main contribution of our learning framework is the agreement-based coupling of the training of multiple models. A model selection approach could still be used to decide which model is used to make predictions. In principle, the consensus method used could simply be an existing model selection method.

\subsubsection{Relationship to Natural Learning}
\label{sec:relationship_to_natural_learning}

The algorithm defined in the previous section is inspired by the ideas presented in section \ref{sec:machine_vs_natural_learning}. Even though a notion of {\em trust} is not established in the algorithm, the consensus prediction step and the augmented loss definition can be thought os as implementing both of the first two steps in the natural learning process described in section \ref{sec:machine_vs_natural_learning}. Furthermore, minimizing the augmented loss function of equation \ref{eq:augmented_loss_function} is equivalent to maximizing agreement with some nature agent, providing known-to-be-true examples, and at the same time maximizing agreement with the other agents (through the consensus loss term). The $\lambda_j$ parameter in that equation corresponds to the trade-off between maximizing these two quantities. Therefore, the algorithm presented in the previous section offers a simple first instance of an agreement-based learning framework inspired by the natural process of learning.

In the next few paragraphs, we describe a handful of potential extensions to our algorithm that can bring us even closer to natural learning, as described earlier.

\paragraph{Trust.} The trust of one agent for others could be explicitly modeled and used to compute {\em personalized consensus} predictions for that agent. There exist various unsupervised and semi-supervised accuracy estimation methods \citep{Platanios:2014, Platanios:2016, Anonymous:logic} that could be used for estimating how much an agent can trust another, about a particular question (e.g, about classifying noun phrases as representing cities or not). Then, the obtained trust estimates could be used to weigh each agent's answer to a question and provide a personalized consensus prediction for an agent to use as training data. This would effectively provide an alternative consensus method that generates different consensus predictions for each agent, based on how much they trust the others.

\paragraph{Decentralized Communication.} In our algorithm we assume that the consensus is formed by combining all model predictions together. This can be prohibitively expensive, in practice, when a large number of models is used. Furthermore, it is also not a realistic representation of natural learning in that humans only consult a few other humans when they have questions. One way to deal with this issue would be to make agents individually able to choose which other agents they want to ask, with respect to a particular question (i.e., these could be the agents they trust the most for a particular kind of question), and what questions they want to ask, more generally. This could be achieved by using an {\em active learning} approach \citep{Settles:2012eg}. \citet{Anonymous:active} propose an approach that could be useful in this setting. This would provide a more realistic, in the sense that it is closer to natural learning, implementation and would result in a much more scalable and easily distributed approach.

There are many more interesting directions to pursue, such as a {\em reinforcement learning} approach (where agreement can be part of a reward function) and a game-theoretic interpretation of this model of natural learning, but these discussions are beyond the scope of this paper.

\section{EXPERIMENTS}
\label{sec:experiments}

In the following sections we describe: (i) the experimental setup, (ii) the datasets we used for our experiments, and (iii) the results we obtained.

\begin{table*}[t]
	\caption{Datasets used in experiments. Number of features corresponds to the dimensionality of the inputs and number of labels corresponds to the dimensionality of the outputs for each dataset. Number of instances corresponds to the total size of the data set. Density corresponds to the proportion of labels that are positive, relative to the total number of labels across all instances.}
	\label{tab:datasets}
	\vspace{-0.5em}
	\begin{sc}
	\small
	\begin{center}
		\begin{tabu} to \linewidth {|[0.8pt]l|[0.6pt]X[r]|X[r]|X[r]|X[r]|[0.8pt]}
			\tabucline[0.8pt]{1-6}
			Dataset				& \#Features		& \#Labels	& \#Instances	& Density (\%) 	\TBstrut \\ \hline
			Delicious			& 501			& 982		& 16,105			& 1.94 			\Tstrut \\
			MediaMill			& 120			& 101		& 43,907			& 4.33 			\\
			RCV1v2  			& 47,236			& 101		& 30,000			& 2.85			\\
			Yahoo-Arts 			& 23,146			& 26 		& 7,484			& 6.36			\\
			Yahoo-Business 		& 21,928			& 27 		& 11,214 		& 6.08			\\
			Yahoo-Computers		& 34,099			& 30 		& 12,444			& 5.60			\\
			Yahoo-Education		& 27,540			& 27 		& 12,030			& 5.42			\\
			Yahoo-Entertainment	& 32,001			& 21 		& 12,730			& 6.73			\\
			Yahoo-Health			& 30,607			& 30 		& 9,205			& 5.99			\\
			Yahoo-Reference		& 39,682			& 30 		& 8,027			& 4.01			\\
			Yahoo-Science		& 37,197			& 30 		& 6,428			& 3.50			\\
			Yahoo-Social			& 52,359			& 30 		& 12,111			& 4.60			\\
			Yahoo-Society		& 31,802			& 27 		& 14,512			& 6.28			\Bstrut \\ \tabucline[0.8pt]{1-6}
		\end{tabu}
	\end{center}\
	\end{sc}
	\vspace{-1em}
\end{table*}

\subsection{EXPERIMENTAL SETUP}
\label{sec:experimental_setup}

For our experiments we consider a multi-label classification setting. This means that the predictions that our models produce are real-valued vectors with values in the interval $[0, 1]$. Each value represents the probability of the corresponding label being equal to $1$. Before each experiment is run, we shuffle the dataset and split it into train and test parts. We run two experiments for each dataset: one using only 5\% of the data to train and one using 50\%. This was done to test our hypothesis that agreement-based learning offers greater benefits when less supervised training data is available. Note that the same seed was used for shuffling the datasets, for all experiments, in order to produce comparable results.

\paragraph{Models.} For each dataset, the set of models that we consider are multi-layer perceptrons (MLPs) with various architectures. They all use leaky rectified linear activation functions (with the leakage parameter set to $0.01$), but we vary the number of hidden layers and number of units per layer to define different architectures, and thus, different models. The architectures used for each dataset are listed in section \ref{sec:datasets}.

\paragraph{Methods.} We run experiments and compare results for the following methods:
\begin{itemize}[noitemsep, topsep=0pt, leftmargin=*]
	\item \uline{CV-5:} This is simply 5-fold leave-one-out cross-validation and forms our baseline. For this method, we perform the following steps during training:
	\begin{enumerate}[noitemsep, topsep=0pt, leftmargin=*]
		\item Split the train dataset into 5 folds.
		\item For each of these folds, train each model on the remaining 4 folds and evaluate its performance on the current fold.
		\item Average these evaluation results over all 5 folds.
		\item Pick the model with the best performance and re-train it using the whole training dataset.
	\end{enumerate}
	The final predictions over the testing dataset are made using the best-performing model, selected in step 4.
	\item \uline{TMV:} Train each model separately and combine their predictions using the trainable majority vote method presented in section \ref{sec:consensus}. Note that this is effectively algorithm \ref{alg:agreement_based_learning} with $K_0\!=\!\infty$ (i.e., keep using the original model loss function during training, without ever switching to the augmented loss function).
	\item \uline{TMV-AL}: Use algorithm \ref{alg:agreement_based_learning} with the trainable majority vote method of section \ref{sec:consensus} as the consensus.
	\item \uline{RBM:} Train each model separately and combine their predictions using the semi-supervised RBM method presented in section \ref{sec:consensus}. Note that this is effectively algorithm \ref{alg:agreement_based_learning} with $K_0\!=\!\infty$ (i.e., keep using the original model loss function during training, without ever switching to the augmented loss function).
	\item \uline{RBM-AL:} Use algorithm \ref{alg:agreement_based_learning} with the semi-supervised RBM method of section \ref{sec:consensus} as the consensus.
\end{itemize}
For both the trainable majority vote and the semi-supervised RBM methods, we use the Adam optimizer \citep{Kingma:2014}. Furthermore, we set a limit of $10,000$ training iterations for the first time the consensus methods are trained, but reduce that to $1,000$ for every time they are re-trained. This reduction is due to the use of warm-starts (i.e., the re-train optimization procedure is initialized at the previous solution) deeming a higher number of iterations unnecessary.

For all methods except cross-validation, we set $N_U=N_L=128$, $K_0=10$, and $K=100$. These parameter values are picked so that our experiments take a reasonable amount of time to run, but varying their values does not seem to affect the results. For the cross-validation method, we still use a batch size of $128$ while training each model separately. Furthermore, we limit all experiments to a maximum number of $2,000$ training iterations in order to limit computation time\footnote{Note that the training procedures for all experiments converged before that limit was reached.}.

\paragraph{Evaluation.} The evaluation metric we use is the macro-averaged area under the precision-recall curve (we refer to this metric as AUC), computed over the testing dataset. This metric is obtained by: (i) computing the area under the precision-recall curve for each output label separately, and (ii) averaging these values. We decided to use this metric as it can measure multi-label classification performance while accounting for the trade-off between precision and recall, for all possible ways in which the probabilistic predictions can be thresholded and converted to boolean values.

\paragraph{Implementation.} We implemented our agreement-based learning framework in Python, using the TensorFlow library \citep{Abadi:2016} as the backend for performing all computations. We have released our implementation as a standalone general purpose library that can be used for training arbitrary TensorFlow models.

\subsection{DATASETS}
\label{sec:datasets}

We use several datasets obtained from the Mulan library website (available at {\small \url{http://mulan.sourceforge.net/datasets-mlc.html}}). Table \ref{tab:datasets} lists the datasets. The {\sc Delicious} dataset contains textual data of web pages as inputs, along with a set corresponding tags as outputs \citep{Tsoumakas:2009}, and is used for automated tag suggestion. It was extracted from the \texttt{del.icio.us} social bookmarking site on the 1\textsuperscript{st} of April, 2007. The {\sc MediaMill} dataset is taken from the MediaMill video indexing challenge \citep{Snoek:2006}. The inputs consist of visual features derived from video sequences and the outputs consist of semantic concepts (e.g., indoors vs. outdoors). The {\sc Yahoo} datasets contain data of web pages that were obtained from Yahoo's top-level categories (e.g., arts, business, etc.) and that are automatically being classified into a number of second-level categories \citep{Ueda:2003}.

The multi-layer perceptron (MLP) architectures used for each dataset are shown in the following list. Please refer to the ``Models'' paragraph of the previous section for information about how these architectures are used in our experiments. The format is a comma-separated list of architectures with each architecture being encoded as a list of numbers. Each number corresponds to the number of units in the corresponding hidden layer. Therefore, \texttt{[32 16]} corresponds to two hidden layers with 32 and 16 units, respectively.

\begin{table*}[t]
	\caption{Area under the precision-recall curve (AUC), computed over the test dataset, for all our experiments. All values are scaled by $10^{-2}$ and each column corresponds to a dataset. There are two sets of experiments: one using 5\% of the data as training data and one using 50\%. Higher AUC values are better.}
	\label{tab:results}
	\vspace{-0.5em}
	\begin{sc}
	\small
	\begin{center}
		\begin{tabu} to \linewidth {|[0.8pt]r@{\,}|[0.6pt]@{\,}X[c]@{\,}|@{\,}X[c]@{\,}|@{\,}X[c]@{\,}|@{\,}X[c]@{\,}|@{\,}X[c]@{\,}|@{\,}X[c]@{\,}|@{\,}X[c]@{\,}|@{\,}X[c]@{\,}|@{\,}X[c]@{\,}|@{\,}X[c]@{\,}|@{\,}X[c]@{\,}|@{\,}X[c]@{\,}|@{\,}X[c]@{\,}|[0.8pt]}
			\tabucline[0.8pt]{1-14}
								& & & & \multicolumn{10}{c|[0.8pt]}{\sc Yahoo} \TBstrut \\ \cline{5-14}
			\multirow{-2}{*}[-3.7em]{$\times 10^{-2}$\phantom{s}} & \side{\sc Delicious} & \side{\sc MediaMill} & \side{\sc RCV1v2} & \side{\sc Arts} & \side{\sc Business} & \side{\sc Computers} & \side{\sc Education} & \side{\sc Entertainment\phantom{s}} & \side{\sc Health} & \side{\sc Reference} & \side{\sc Science} & \side{\sc Social} & \side{\sc Society} \\ & & & & & & & & & & & & & \TBstrut \\ \tabucline[0.6pt]{1-14}
			\multicolumn{14}{|[0.8pt]c|[0.8pt]}{5\% Train Data / 95\% Test Data} \TBstrut \\ \tabucline[0.6pt]{1-14}
			CV-5		& $11.09$ & $15.22$ & $52.30$ & $13.36$ & $14.87$ & $14.25$ & $13.30$ & $22.37$ & $15.87$ & $\phantom{0}9.47$ & $10.79$ & $12.85$ & $16.06$ \Tstrut \\
			TMV		& $11.35$ & $16.69$ & $46.20$ & $16.31$ & $12.24$ & $16.08$ & $18.80$ & $29.44$ & $20.06$ & $23.42$ & $15.64$ & $21.87$ & $19.91$ \\
			TMV-AL	& $12.01$ & $20.19$ & $53.39$ & $12.04$ & $\phantom{0}7.29$ & $\phantom{0}9.08$ & $11.33$ & $22.20$ & $12.31$ & $\phantom{0}7.82$ & $\phantom{0}8.16$ & $\phantom{0}8.97$ & $14.54$ \\ 
			RBM		& $12.66$ & $28.02$ & $55.63$ & $26.93$ & $34.89$ & $30.08$ & $34.41$ & $37.88$ & $42.48$ & $41.96$ & $29.16$ & $30.29$ & $30.51$ \\
			RBM-AL	& \best{$23.30$} & \best{$39.86$} & \best{$59.87$} & \best{$38.26$} & \best{$49.07$} & \best{$49.86$} & \best{$46.48$} & \best{$39.12$} & \best{$50.65$} & \best{$44.63$} & \best{$47.37$} & \best{$49.55$} & \best{$30.61$} \Bstrut \\ \tabucline[0.6pt]{1-14}
			\multicolumn{14}{|[0.8pt]c|[0.8pt]}{50\% Train Data / 50\% Test Data} \TBstrut \\ \tabucline[0.6pt]{1-14}
			CV-5	& $14.45$ & $16.41$ & $94.28$ & $31.54$ & $32.42$ & $40.07$ & $32.54$ & $43.02$ & $43.77$ & $25.75$ & $34.48$ & $44.81$ & $34.96$ \Tstrut \\
			TMV		& $12.84$ & $22.55$ & $89.56$ & $29.29$ & $30.86$ & $32.85$ & $27.05$ & $43.77$ & $36.93$ & $30.08$ & $29.41$ & $33.51$ & $31.29$ \\
			TMV-AL	& $13.35$ & $20.38$ & $92.34$ & $25.13$ & $21.02$ & $27.10$ & $25.86$ & $34.63$ & $24.30$ & $22.43$ & $21.63$ & $31.15$ & $30.97$ \\
			RBM		& $28.64$ & $41.00$ & $96.43$ & $39.89$ & $40.00$ & $48.92$ & $43.58$ & $54.32$ & $49.82$ & $30.72$ & $41.00$ & $38.32$ & $39.44$ \\
			RBM-AL	& \best{$42.21$} & \best{$46.99$} & \best{$98.87$} & \best{$41.52$} & \best{$49.42$} & \best{$50.30$} & \best{$47.72$} & \best{$57.27$} & \best{$53.76$} & \best{$41.53$} & \best{$48.86$} & \best{$51.23$} & \best{$44.45$} \Bstrut \\ \tabucline[0.8pt]{1-14}
		\end{tabu}
	\end{center}
	\end{sc}
	\vspace{-1em}
\end{table*}

\begin{flushleft}
\begin{enumerate}[noitemsep, topsep=0pt, leftmargin=*]
	\item \uline{Delicious:} \texttt{\small [16]}, \texttt{\small [256]}, \texttt{\small [16~16]}, \texttt{\small [256~256]}, \texttt{\small [512~256]}, \texttt{\small [1024~512~256]}, \texttt{\small [16~16~16~16]}, \texttt{\small [256~256~256~256].}
	\item \uline{MediaMill:} \texttt{\small [1]}, \texttt{\small [8]}, \texttt{\small [16~8]}, \texttt{\small [32~16]}, \texttt{\small [256~128]}, \texttt{\small [1024~1024]}, \texttt{\small [2048~2048]}, \texttt{\small [128~32~8]}, \texttt{\small [128~64~32~16].}
	\item \uline{RCV1v2 and Yahoo (all datasets):} \texttt{\small [1]}, \texttt{\small [8]}, \texttt{\small [16~8]}, \texttt{\small [32~16]}, \texttt{\small [256~128]}, \texttt{\small [128~32~8]}, \texttt{\small [128~64~32~16].}
\end{enumerate}
\end{flushleft}
Note that these architectures were chosen arbitrarily in order to have both very simple and very complex models that can overfit easily.

\subsection{RESULTS}
\label{sec:results}

The results from our experiments are shown in table \ref{tab:results}. Our first observation is that the {\em RBM-AL method always outperforms all of the competing methods}. This is a very strong result for our proposed agreement-based learning framework. However, it is interesting to also note that TMV-AL does not always outperform TMV. In fact, for the experiments that we ran, it almost always gets outperformed by TMV. The most likely reason for that is that the consensus prediction that TMV generates is not very accurate and is also potentially too sensitive to dependencies among the models\footnote{A highly dependent scenario would be one where we have one model and nine copies of another model. Such a scenario could be detrimental to a bad consensus method.}. Thus, a necessary condition for our proposed algorithm to work is that the consensus method used performs well. This condition seems to be satisfied by the RBM consensus. It would also probably be satisfied by other state-of-the-art ensemble methods \citep{Platanios:2014, Platanios:2016, Anonymous:logic} that use similar intuition for the dependencies among the models. On a related note, we, more generally, expect that an agreement-based learning framework would always work better when the models being used are not highly dependent.

Furthermore, we observe that, even though RBM-AL always outperforms RBM, which is the second best-performing method in our experiments, it does so by a larger margin when more unlabeled data are used (i.e., 95\% test data vs. 50\% test data). This observation agrees with our expectation of section \ref{sec:coupling_training}.

Finally, it is interesting to point out that RBM-AL always outperforms cross-validation. Perhaps even more interestingly, using RBM-AL with 5\% of the dataset as training data outperforms using cross-validation with 50\% of the dataset as training data. This is a significant result that makes agreement-based learning seem like a very strong alternative to cross-validation.

\section{CONCLUSION}

In this paper, we have proposed an agreement-based learning framework that prevents many of the pitfalls associated with model selection. This framework is inspired by human learning and it relies on coupling the training of multiple models by encouraging them to agree on their predictions while training. We also proposed an algorithm defined within this framework which was shown to significantly outperform alternative methods in practice, and whose performance was shown to improve further with the availability of unlabeled data. There exist many potential future directions for this work. We intend to pursue the directions related to trust and decentralized communication that were discussed in section \ref{sec:relationship_to_natural_learning}.

%


\bibliography{paper}
\bibliographystyle{icml2017}

\end{document}